\PassOptionsToPackage{table}{xcolor}
\documentclass[11pt]{article}
\usepackage[preprint]{acl}
\usepackage[utf8]{inputenc}
\usepackage[T1]{fontenc}
\usepackage{microtype}
\usepackage{inconsolata}
\usepackage{graphicx}
\usepackage{booktabs}
\usepackage{multirow}
\usepackage{makecell}

\usepackage{times}
\usepackage[table]{xcolor}
\usepackage{pifont}

\newcommand{\rwcheck}{\textcolor{green!55!black}{\ding{51}}}
\newcommand{\rwxmark}{\textcolor{red!70!black}{\ding{55}}}
\usepackage{array}
\usepackage{tabularx}
\usepackage{amsmath,amssymb,mathtools}
\usepackage{algorithm}
\usepackage{algpseudocode}
\usepackage{xcolor}
\usepackage[most]{tcolorbox}
\usepackage{url}
\usepackage{hyperref}
\usepackage[capitalize,noabbrev]{cleveref}
\usepackage{pifont}

\newcommand{\method}{\textsc{MegaMem}}

\newtcolorbox{promptbox}[2][blue]{%
  colback=#1!4,
  colframe=#1!55!black,
  colbacktitle=#1!55!black,
  coltitle=white,
  boxrule=0.5pt,
  arc=1pt,
  left=4pt,right=4pt,top=4pt,bottom=4pt,
  title=\textbf{#2},
  fonttitle=\small,
  fontupper=\footnotesize
}
\makeatletter
\renewcommand{\theALG@line}{\thealgorithm.\arabic{ALG@line}}
\renewcommand{\theHALG@line}{\thealgorithm.\arabic{ALG@line}}
\makeatother

\title{\method{}: A Retrieval Solution for Ultra-Large Context Windows}
\author{
{\normalfont Xinyuan Song$^{1,2}$ \quad
Bowen Zhu$^{1}$} \quad
{\normalfont Hasibul Haque$^{1}$ \quad
Liang Zhao$^{1,2}$} \\
{\normalfont $^{1}$Causal Dynamics Lab, USA \quad
$^{2}$Emory University, USA \quad
}
}

\begin{document}
\maketitle

\begin{abstract}
Modern language models and agents increasingly require persistent memory for complete codebases, long interaction histories, and heterogeneous enterprise records. The key challenge is to keep hundreds of millions of tokens searchable while passing only bounded source evidence to the answer model. We introduce \method{}, a source-resolved dual-view retrieval system that separates semantic access from generation evidence. Distilled records and detailed evidence are searched with original and transformed queries; every distilled hit resolves to an immutable source ID before reciprocal-rank fusion, deduplication, and cross-encoder reranking; and only the highest-ranked detailed evidence within a fixed budget supports generation. Post-answer attribution then identifies which loaded sources support the fixed answer. We evaluate \method{} on EnterpriseRAG-Bench, which contains more than 500,000 heterogeneous enterprise documents and approximately 650M tokens. \method{} improves Overall from 68.22 to 82.26 and reaches 86.50 Correctness. These results show that \method{} supports ultra-large persistent memory while preserving strong answer accuracy under a bounded generation context. By separating searchable memory scale from answer-context size, \method{} provides a practical path toward accurate retrieval over memories ranging from hundreds of millions to one billion tokens. Our code is available at \url{https://github.com/xfab-xinyuansong/MegaMem.git}.
\end{abstract}

\section{Introduction}

\label{sec:introduction}

Modern large language models and agents increasingly depend on large context to support long-term interaction, planning, tool use, and knowledge-intensive decision making~\cite{packer2023memgpt,wang2023longmem,zhong2023memorybank,maharana2024locomo,wu2025longmemeval,chhikara2025mem0,kang2025memoryos}. Memory management has therefore become a core problem in agent systems, since an agent must retain observations, task states, interaction histories, and prior experience across steps and sessions. In practice, modern language systems require persistent memory that can store complete codebases, long interaction histories, and heterogeneous enterprise records~\cite{sun2026enterpriseragbench,choubey2025deepsearch,yu2025ekrag}. Native long-context models can process millions of tokens in one call~\cite{reid2024gemini15}, while retrieval-based models can search datastores containing trillions of tokens~\cite{borgeaud2022retro}. Persistent-memory systems instead store information outside the active context and retrieve selected content when needed~\cite{packer2023memgpt,wang2023longmem}. Recent systems further organize, update, and retrieve long-term memory to support continued interaction, personalization, and cross-session reasoning~\cite{zhong2023memorybank,maharana2024locomo,wu2025longmemeval,chhikara2025mem0,kang2025memoryos}.

However, only a small fraction of such memory usually supports one answer. Directly loading more content creates a quality--cost trade-off: relevant evidence may be missed because of its position~\cite{liu2023lostinmiddle}, performance can decline as context length and document count increase~\cite{hsieh2024ruler,levy2025moredocuments,du2025contextlength}, and irrelevant passages can distract the model~\cite{cuconasu2024powerofnoise}. Processing the full repository also causes attention computation, latency, and serving cost to grow with stored history. Existing methods address this problem through hierarchical summaries~\cite{sarthi2024raptor}, graph-based corpus representations~\cite{edge2024graphrag,guo2024lightrag,gutierrez2024hipporag}, global memory and retrieval clues~\cite{qian2024memorag}, or external memory management~\cite{packer2023memgpt,wang2023longmem,chhikara2025mem0,kang2025memoryos}. These methods improve access to large stores, but compressed records may omit dates, exceptions, conditions, or conflicts needed for a correct answer. In contrast, retrieving only detailed chunks preserves exact support but increases distractor exposure and weakens retrieval precision as the corpus grows~\cite{cuconasu2024powerofnoise,levy2024sametaskmoretokens,levy2025moredocuments}.

Ultra-large memory retrieval must therefore support three goals: semantic retrieval across different expressions, source-faithful evidence for generation, and precise attribution to the sources used. We distinguish \emph{persistent context}, the full corpus that the system can search, from \emph{evidence context}, the bounded detailed source evidence consumed for one answer. The goal is to scale the former without increasing the latter at the same rate. We introduce \method{}, which assigns these goals to separate stages: distilled records support retrieval, each distilled hit resolves to an immutable detailed-source identifier, only detailed source evidence enters generation, and post-answer attribution identifies the loaded sources that support the fixed answer~\cite{hsia2025ragged,levy2024sametaskmoretokens}.

\begin{figure*}[!ht]

\centering
\includegraphics[width=0.95\textwidth]{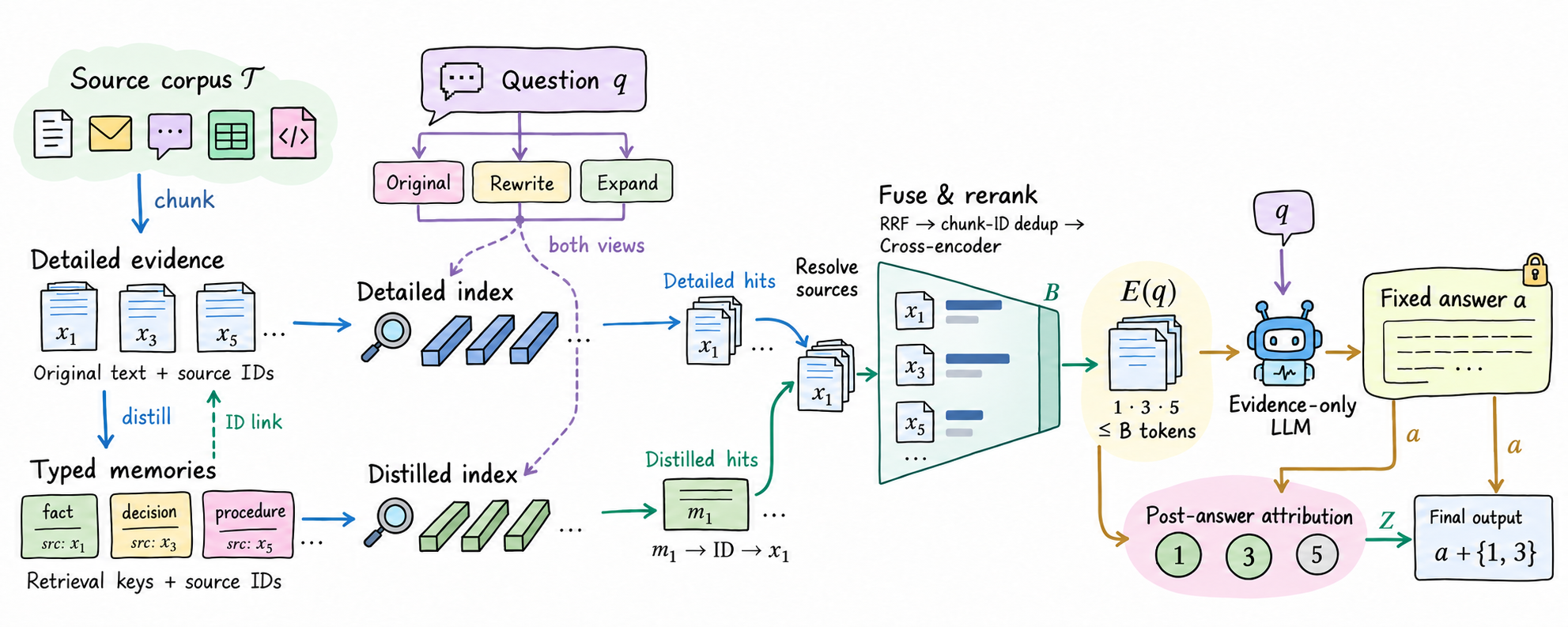}

\caption{\textbf{Overview of MegaMem.} Dual-view retrieval combines detailed evidence with distilled typed memories, resolves all retrieved memories to their original sources, and reranks evidence under a fixed context budget. The LLM generates an evidence-grounded answer, followed by source attribution.}
\label{fig:overview}

\end{figure*}

Ultra-large memory retrieval must support semantic access, source-faithful generation, and precise attribution. We distinguish \emph{persistent context}, the full searchable corpus, from \emph{evidence context}, the bounded detailed source evidence used for one answer. \method{} scales the former while keeping the latter fixed: distilled records support retrieval, every distilled hit resolves to detailed source evidence through an immutable source ID, and only the highest-ranked resolved chunks enter generation. Original and transformed queries search both views, followed by reciprocal-rank fusion, deduplication, and cross-encoder reranking. As shown in Figure~\ref{fig:overview}, this design separates persistent memory, retrieval candidates, loaded evidence, and reported sources, allowing \method{} to operate over hundreds of millions of tokens without passing the full memory to the answer model.

This design builds on prior work in native long context, persistent memory, structured retrieval, multi-stage retrieval, and attribution~\cite{reid2024gemini15,packer2023memgpt,sarthi2024raptor,cormack2009rrf,gao2023enabling}, and leads to three research questions. First, does separating retrieval records from generation evidence improve end-to-end answer quality? Second, which components---dual-view retrieval, distillation, query transformation, reranking, and post-answer attribution---produce the observed gains? Third, as persistent memory grows under a fixed evidence budget, does performance decline because retrieval fails to find the required evidence or because the answer model cannot use the selected evidence?

We evaluate these questions on EnterpriseRAG-Bench, a recent benchmark for retrieval over realistic enterprise knowledge~\cite{sun2026enterpriseragbench}. It contains more than 500,000 documents from heterogeneous enterprise sources and 500 questions covering retrieval, multi-document reasoning, conflict resolution, constrained search, and missing-information cases. The full corpus used in our study contains approximately 650M tokens, providing a direct test of retrieval over ultra-large and noisy memory. On the disjoint 400-question validation split, \method{} provides an effective solution to ultra-large memory retrieval, improving Overall from 68.22 to 82.26 and reaching 86.50 Correctness. The strongest gains come from dual-view retrieval and distillation, confirming that separating semantic access from source-faithful evidence is central to the design. Post-answer attribution further reduces the reported source set without changing answer quality, improving source reporting while preserving the final prediction. This work makes three contributions:
\begin{itemize}
    \item We study the problem of ultra-large persistent memory at the scale of 250M to 1B tokens, where a language system must keep the full memory searchable while passing only a bounded amount of evidence to the answer model.
    \item We introduce \method{}, a source-resolved dual-view retrieval system that searches both distilled records and detailed evidence with original and transformed queries, resolves all distilled hits to immutable source IDs before reciprocal-rank fusion, deduplication, and cross-encoder reranking, loads only the highest-ranked detailed evidence under a fixed budget, and applies post-answer attribution to the fixed answer.
    \item We evaluate \method{} against strong retrieval and memory baselines and provide controlled analyses of end-to-end quality, component effects, wall-clock cost, token use, evidence-budget trade-offs, and memory scaling. \method{} achieves the best overall performance while keeping the generation context bounded as persistent memory grows.
\end{itemize}

\section{Related Work}
\label{app:related-work}

\paragraph{Long context and persistent memory.}
Million-token context windows do not guarantee reliable long-context utilization, which remains sensitive to length, position, and distractors~\citep{reid2024gemini15,zhang2024infinitebench,hsieh2024ruler,bai2024longbenchv2,liu2023lostinmiddle,levy2024sametaskmoretokens}. Retrieval therefore remains useful alongside long context~\citep{lewis2020rag,jiang2024longrag,xu2024retrievalmeets}, while persistent-memory systems extend external memory across interactions~\citep{packer2023memgpt,wu2025longmemeval,chhikara2025mem0,kang2025memoryos}. \method{} targets large persistent corpora by separating searchable memory from bounded generation evidence.

\paragraph{Structured retrieval and evidence selection.}
Prior work structures retrieval through hierarchical summaries, graphs, and memory-generated clues~\citep{sarthi2024raptor,edge2024graphrag,guo2024lightrag,gutierrez2024hipporag,qian2024memorag,jin2025longrefiner}, while query transformation, heterogeneous retrievers, and reranking improve candidate discovery~\citep{karpukhin2020dpr,formal2021spladev2,santhanam2022colbertv2,wang2023query2doc,liu2025exp4fuse,nogueira2019passage}. \method{} combines distilled and detailed retrieval views, resolves compressed memories to source evidence before generation, and separates evidence selection from post-answer attribution~\citep{gao2023enabling,qi2024mirage}.

\paragraph{Enterprise retrieval.}
Enterprise retrieval must handle heterogeneous sources, private terminology, and answerability at scale. EKRAG, Deep Search, and EnterpriseRAG-Bench study complementary aspects of this setting~\citep{yu2025ekrag,choubey2025deepsearch,sun2026enterpriseragbench}. We use EnterpriseRAG-Bench as our primary benchmark and FinanceBench, HotpotQA, LoCoMo, and UltraDomain to evaluate transfer across retrieval settings.

\paragraph{How MegaMem Differs from Prior Retrieval and Memory Systems}
\begin{table*}[!ht]
\centering
\caption{\textbf{Comparison with closely related retrieval and memory systems.} We compare key capabilities across structured memory, source-resolved retrieval, evidence construction, attribution, and scalability.}
\label{tab:related-work-comparison}

\begingroup
\small
\setlength{\tabcolsep}{2pt}

\resizebox{0.8\textwidth}{!}{%
\begin{tabular}{
@{}
>{\raggedright\arraybackslash}p{0.31\textwidth}
>{\centering\arraybackslash}p{0.09\textwidth}
>{\centering\arraybackslash}p{0.09\textwidth}
>{\centering\arraybackslash}p{0.09\textwidth}
>{\centering\arraybackslash}p{0.09\textwidth}
>{\centering\arraybackslash}p{0.09\textwidth}
>{\centering\arraybackslash}p{0.09\textwidth}
>{\centering\arraybackslash}p{0.09\textwidth}
@{}
}

\toprule
\textbf{Method}
& \textbf{Structured / Compressed Memory}
& \textbf{Direct Source Resolution}
& \textbf{Multiple Retrieval Views}
& \textbf{Original-Source Generation}
& \textbf{Bounded Evidence}
& \textbf{Post-hoc Attribution}
& \textbf{Corpus-Growth Analysis} \\
\midrule

Dense X Retrieval \citep{chen2024densex}
& \rwcheck & \rwcheck & \rwxmark & \rwcheck & \rwxmark & \rwxmark & \rwxmark \\

Fidelity Before Structure \citep{an2026cogcanvas}
& \rwcheck & \rwcheck & \rwcheck & \rwcheck & \rwxmark & \rwxmark & \rwxmark \\

MemoRAG \citep{qian2024memorag}
& \rwcheck & \rwxmark & \rwcheck & \rwcheck & \rwxmark & \rwxmark & \rwxmark \\

HippoRAG 2 \citep{gutierrez2025hipporag2}
& \rwcheck & \rwcheck & \rwcheck & \rwcheck & \rwxmark & \rwxmark & \rwxmark \\

FlexStructRAG \citep{chen2026flexstructrag}
& \rwcheck & \rwcheck & \rwcheck & \rwcheck & \rwxmark & \rwxmark & \rwxmark \\

EverMemOS \citep{hu2026evermemos}
& \rwcheck & \rwxmark & \rwcheck & \rwxmark & \rwcheck & \rwxmark & \rwxmark \\

LongRefiner \citep{jin2025longrefiner}
& \rwxmark & \rwcheck & \rwxmark & \rwcheck & \rwcheck & \rwxmark & \rwxmark \\

Exp4Fuse \citep{liu2025exp4fuse}
& \rwxmark & \rwxmark & \rwcheck & \rwcheck & \rwcheck & \rwxmark & \rwxmark \\

MIRAGE; ContextCite \citep{qi2024mirage,cohenwang2024contextcite}
& \rwxmark & \rwcheck & \rwxmark & \rwcheck & \rwcheck & \rwcheck & \rwxmark \\

RAGGED \citep{hsia2025ragged}
& \rwxmark & \rwcheck & \rwcheck & \rwcheck & \rwcheck & \rwxmark & \rwcheck \\

\midrule
\rowcolor{blue!9}
\textbf{\method{}}
& \rwcheck & \rwcheck & \rwcheck & \rwcheck & \rwcheck & \rwcheck & \rwcheck \\

\bottomrule
\end{tabular}%
}

\endgroup
\end{table*}

Table~\ref{tab:related-work-comparison} positions \method{} against closely related retrieval, memory, and attribution systems across the key capabilities relevant to our design. While individual components have precedents, \method{} combines structured memory, source-resolved multi-view retrieval, bounded original-source evidence, and post-hoc attribution within a unified pipeline.

\section{Problem Formulation}

\label{sec:formulation}

\subsection{Persistent Context and Evidence Context}

Let \(\mathcal{D}\) denote the full persistent corpus with \(T\) tokens. For a question \(q\), the system retrieves an evidence set \(E(q)\subseteq\mathcal{D}\) under a fixed budget \(B\):
\begin{equation}
E(q)\subseteq\mathcal{D}, \qquad \operatorname{tok}(E(q))\leq B, \qquad B\ll T.
\label{eq:bounded-evidence}
\end{equation}
We refer to \(\mathcal{D}\) as the \emph{persistent context} and \(E(q)\) as the \emph{evidence context}. The goal is to scale the searchable corpus while keeping the evidence context bounded, even though retrieval becomes harder as distractors increase.

\subsection{Source-Resolved Retrieval}
\label{sec:resolution-invariant}

Prior work uses compressed or structured representations to improve retrieval over large corpora~\cite{sarthi2024raptor,edge2024graphrag,qian2024memorag}. In \method{}, such representations are used only to locate relevant content: every compressed hit is resolved to its corresponding detailed source chunk before fusion, reranking, and generation. The answer is therefore produced only from detailed evidence:
\begin{equation}
a=\mathcal{A}\bigl(q,E(q)\bigr), \qquad E(q)\subseteq\mathcal{X},
\label{eq:source-resolved-answer}
\end{equation}
where \(\mathcal{X}\) is the set of detailed source chunks.

\section{Method}
\label{sec:method}

\method{} consists of three stages: dual-view memory construction, multi-route retrieval, and bounded-evidence generation with post-answer attribution. As illustrated in Figure~\ref{fig:overview}, the system separates retrieval-oriented representations from generation evidence. Distilled memories improve semantic access to the corpus, but every retrieved memory is resolved to its corresponding detailed source chunk before fusion, reranking, and context construction. Consequently, only detailed source evidence is passed to the answer model or exposed as a reported source.

\subsection{Dual-View Memory Construction}
\label{sec:offline}

Given a document collection, we first normalize each document and segment it into section-aware evidence chunks using headings and paragraph boundaries. Long paragraphs are further split by sentence and token boundaries. This produces a detailed evidence view
\(\mathcal{X}=\{x_1,\ldots,x_n\}\), where each chunk retains its original content and source metadata.

For each detailed chunk \(x_i\), an extractor produces up to three compact typed memories:
\begin{equation}
m_{ij}=(\tau_{ij},u_{ij},v_{ij},\pi_{ij}), \qquad \pi_{ij}=i,
\label{eq:typed-memory}
\end{equation}
where \(\tau_{ij}\) specifies the memory type, \(u_{ij}\) is a retrieval key, \(v_{ij}\) is a concise statement supported by \(x_i\), and \(\pi_{ij}\) links the memory to its source chunk. We use five memory types: fact, procedure, definition, requirement, and decision. Chunks without useful semantic content may produce no memory.

The resulting corpus is indexed in two complementary views. The detailed index stores embeddings of the original evidence chunks together with their source metadata, while the distilled index stores embeddings of the typed memories and their source pointers. The distilled view is optimized for semantic matching, whereas the detailed view preserves the exact wording, conditions, and provenance required for answering. Both views therefore refer to the same underlying evidence rather than forming separate evidence pools.

Atomic memories are extracted with \texttt{gpt-5.4-mini}~\citep{openai2026gpt54mini}. \emph{MiniExtractor} retains only these atomic memories, whereas \emph{FullExtractor} additionally uses \texttt{gpt-5.4} to construct higher-level retrieval abstractions~\citep{openai2026gpt54}. Both configurations use \texttt{text-embedding-3-small} for indexing~\citep{openai2024embedding3}. 


The hierarchy and relation structures explored during development are not used in the final system. As shown in Table~\ref{tab:development}, the final design retains only the detailed and distilled views. Algorithm~\ref{alg:build} in Appendix~\ref{app:implementation} provides the complete construction procedure.

\subsection{Multi-Route Retrieval and Evidence Resolution}
\label{sec:retrieval}

Given a question \(q\), \method{} constructs multiple query routes from the original question, including a canonical rewrite and terminology-diverse expansions. Each route searches both the detailed index \(I_d\) and the distilled index \(I_m\). The detailed route directly retrieves source chunks, whereas the distilled route retrieves compact memories that may better match alternative expressions of the same information need.

All distilled hits are resolved to their corresponding detailed chunks before candidate aggregation. Let \(\operatorname{Retr}(r,I)\) denote the ranked set of top-\(k\) items retrieved from index \(I\) using query route \(r\), and let \(\pi(m)\) map a distilled memory \(m\) to its associated detailed source chunk. The candidate set is therefore defined over detailed evidence identities:
\begin{equation}
C(q)=\bigcup_{r\in\mathcal{Q}(q)}
\left(
\operatorname{Retr}(r,I_d)
\cup
\pi\left(\operatorname{Retr}(r,I_m)\right)
\right),
\label{eq:candidate-set}
\end{equation}
where \(\mathcal{Q}(q)\) contains the original and transformed queries, \(I_d\) is the detailed index, and \(I_m\) is the distilled index. This resolution step allows multiple retrieval routes to improve recall without treating compressed memories as independent evidence items.

Because scores from different routes are not directly comparable, we combine their rankings using weighted reciprocal rank fusion~\citep{cormack2009rrf}:
\begin{equation}
S_{\mathrm{RRF}}(x\mid q)=\sum_{r\in\mathcal{L}(q)}\frac{w_r\,\mathbf{1}[x\in r]}{\gamma+\operatorname{rank}_{r}(x)},
\label{eq:rrf}
\end{equation}
where \(\mathcal{L}(q)\) is the set of ranked retrieval lists, \(w_r\) is the weight assigned to route \(r\), and \(\gamma=60\). We assign larger weights to the original and canonical semantic routes than to auxiliary expansions.

A cross-encoder then reranks the fused candidate pool using the question and full detailed chunk content~\citep{nogueira2019passage,ma2023rankllama}. Duplicate candidates are merged by source identity, and the highest-ranked chunks are packed until the evidence budget \(B\) is reached. The resulting evidence context contains only detailed source chunks, while query transformation and distilled retrieval serve solely to improve candidate discovery.

\begin{algorithm}[!ht]
\caption{Multi-route retrieval with source resolution}
\label{alg:retrieve}
\small
\begin{algorithmic}[1]
\Require question \(q\), detailed index \(I_d\), distilled index \(I_m\), evidence budget \(B\)
\State \(\mathcal{Q}\gets\Call{Transform}{q}\cup\{q\}\)
\State retrieve ranked candidates from \(I_d\) and \(I_m\) for each query in \(\mathcal{Q}\)
\State resolve every distilled candidate \(m\) to its detailed source chunk \(\pi(m)\)
\State fuse resolved rankings with weighted reciprocal rank fusion
\State rerank and deduplicate candidates by detailed source identity
\State \(E(q)\gets\Call{Pack}{\text{ranked detailed chunks},B}\)
\State \Return \(E(q)\)
\end{algorithmic}
\end{algorithm}

\subsection{Bounded Generation and Post-Answer Attribution}
\label{sec:evidence}

The retrieval stage returns a bounded set of detailed evidence chunks \(E(q)\), which forms the complete context available to the answer model. The answer is generated under an evidence-only instruction:
\begin{equation}
a=\mathcal{A}\bigl(q,E(q)\bigr).
\label{eq:bounded-generation}
\end{equation}
No distilled memory or unselected source is exposed during generation.

After the answer is fixed, a separate attribution module identifies which retrieved documents directly support its claims:
\begin{equation}
Z=\mathcal{C}\bigl(a,E(q)\bigr), \qquad Z\subseteq\operatorname{docs}\bigl(E(q)\bigr).
\label{eq:post-answer-attribution}
\end{equation}
Because attribution is performed after generation, it can only remove unsupported or unused sources from the reported set; it cannot revise the answer or introduce evidence that was not shown to the answer model.

This ordering separates three roles in the pipeline. Query transformation broadens candidate recall, reranking selects the evidence used for generation, and post-answer attribution reduces the final source set. In our ablation, removing attribution leaves answer quality unchanged but increases the average number of reported documents from \(2.31\) to \(5.00\), confirming that the module primarily improves source precision rather than answer correctness. The complete runtime procedure is provided in Algorithm~\ref{alg:answer} in Appendix~\ref{app:implementation}~\citep{gao2023enabling,qi2024mirage}.

\section{Experimental Setup}
\label{sec:experiments}

\paragraph{Benchmarks and protocols.}
We primarily evaluate \method{} on EnterpriseRAG-Bench, which contains 500 questions over more than 500,000 heterogeneous enterprise documents~\citep{sun2026enterpriseragbench}. We use 100 questions for architecture, prompt, and hyperparameter development, then freeze the system and evaluate on the disjoint 400-question validation split. Table~\ref{tab:protocol-main} summarizes the main analyses. Because the question sets, memory scales, and extractors differ across protocols, results should be compared only within matched settings. We additionally evaluate transfer on FinanceBench, HotpotQA, LoCoMo, and UltraDomain~\citep{islam2023financebench,yang2018hotpotqa,maharana2024locomo,qian2024memorag}.

\begin{table*}[!ht]
\centering
\caption{\textbf{Main EnterpriseRAG-Bench protocols.} Results are comparable only within matched settings.}

\label{tab:protocol-main}
\resizebox{0.85\linewidth}{!}{
\begin{tabular}{@{}l l l l l@{}}
\toprule
Analysis & Questions & Memory scale & Extractor & Purpose \\
\midrule
Headline comparison & Validation, 400 & 10M & FullExtractor & End-to-end quality \\
Scaling trace & Validation, 400 & 20M--250M & MiniExtractor & Scale behavior \\
Component ablation & Full benchmark, 500 & 10M & FullExtractor & Component diagnosis \\
Gold intervention & Validation, 400 & 20M/60M/250M & Both & Failure localization \\
\bottomrule
\end{tabular}
}

\end{table*}

\paragraph{Baselines and metrics.}
We compare against lexical and dense retrieval baselines, including BM25, DPR, and Contriever~\citep{robertson2009bm25,karpukhin2020dpr,izacard2022contriever}, as well as hierarchical, graph-based, and memory-guided retrieval methods~\citep{sarthi2024raptor,edge2024graphrag,guo2024lightrag,qian2024memorag}. We adopt the evaluation metrics defined in EnterpriseRAG-Bench~\citep{sun2026enterpriseragbench}. \emph{Correctness} measures whether the generated answer reaches the correct conclusion while preserving the required scope, conditions, and conflicts. \emph{Completeness} measures how fully the answer covers the facts required by the reference answer. \emph{Overall} is the benchmark's aggregate answer-quality score derived from Correctness and Completeness. \emph{Document Recall} measures the fraction of gold source documents recovered by the system. \emph{InvDocs} measures the fraction of reported documents judged invalid or unsupported, with lower values indicating better source precision. Atomic extraction uses \texttt{gpt-5.4-mini}~\citep{openai2026gpt54mini}; higher-level abstraction, query transformation, answering, attribution, and evaluation use \texttt{gpt-5.4}~\citep{openai2026gpt54}; and all dense indices use \texttt{text-embedding-3-small}~\citep{openai2024embedding3}. Appendix~\ref{app:protocol} provides the full protocol and implementation details.

\section{Results}

\label{sec:results}

We first present the main scaling results, showing that \method{} remains effective as persistent memory grows to hundreds of millions of tokens under a fixed evidence budget. We then compare \method{} with direct retrieval baselines and analyze the contribution of individual components. Together, these experiments demonstrate that \method{} provides a strong and scalable solution for retrieval over ultra-large persistent memory. Because the protocols differ in question set, memory scale, and extractor, we interpret each result within its own experimental setting.

\subsection{Scaling to Ultra-Large Persistent Memory}
\label{sec:scaling}

\begin{figure}[!ht]

\centering
\includegraphics[width=\columnwidth]{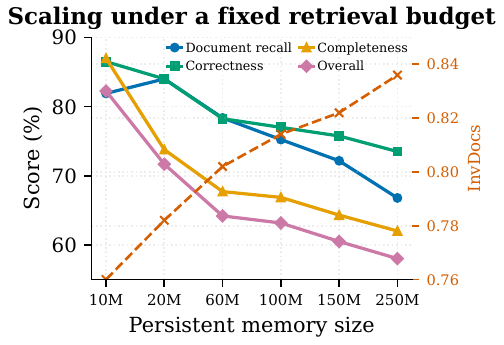}

\caption{\textbf{Performance as persistent memory grows under a fixed evidence budget.} \method{} remains operational from 20M to 250M tokens without increasing the context provided to the answer model.}
\label{fig:scaling}

\end{figure}

We next evaluate whether \method{} remains effective as the persistent corpus grows from 20M to 250M tokens while the retrieval and evidence budgets remain fixed. As shown in \Cref{tab:scaling,fig:scaling}, \method{} continues to produce useful answers throughout this range, achieving an Overall score of 58.02 and a Correctness score of 73.50 even at 250M tokens. This demonstrates that the system can keep hundreds of millions of tokens searchable without increasing the generation context.

\begin{table*}[!ht]
\centering
\caption{\textbf{EnterpriseRAG-Bench performance across persistent-memory scales.} All rows use the 400-question validation split, MiniExtractor, GPT-5.4 answering, and fixed retrieval and evidence budgets. InvDocs is in \([0,1]\), and lower is better.}

\label{tab:scaling}
\resizebox{0.8\textwidth}{!}{%
\begin{tabular}{@{}r r r r r r@{}}
\toprule
Memory scale & Overall & Document Recall & Correctness & Completeness & InvDocs \\
\midrule
20M  & 71.67 & 84.00 & 84.00 & 73.78 & 0.782 \\
60M  & 64.19 & 78.36 & 78.25 & 67.71 & 0.802 \\
100M & 63.17 & 75.23 & 77.00 & 66.87 & 0.814 \\
150M & 60.49 & 72.16 & 75.75 & 64.31 & 0.822 \\
250M & 58.02 & 66.79 & 73.50 & 62.01 & 0.836 \\
\bottomrule
\end{tabular}%
}

\end{table*}

Although retrieval becomes more difficult as the corpus grows, performance degrades gradually rather than collapsing. From 20M to 250M tokens, Correctness remains above 73\%, while Document Recall remains above 66\%. These results support the central claim of \method{}: ultra-large persistent memory can be made addressable while keeping the evidence context bounded.

\subsection{Comparison with Direct Retrieval Baselines}
\label{sec:headline}

Table~\ref{tab:headline} compares \method{} with standard retrieval pipelines that directly retrieve evidence from the original corpus, including lexical and single-vector dense RAG~\citep{robertson2009bm25,karpukhin2020dpr,izacard2022contriever}. We also include compressed-memory variants, the published EnterpriseRAG-Bench leader, and intermediate \method{} configurations. All systems are evaluated on the same 400-question validation split at the 10M memory scale.

\method{} achieves the best Overall, Correctness, Completeness, and Document Recall. Relative to the published benchmark leader, the final system improves Overall by \(20.58\%\), Correctness by \(6.00\%\), Completeness by \(19.38\%\), and Document Recall by \(3.64\%\). These gains show that combining distilled retrieval keys with source-resolved detailed evidence is more effective than directly retrieving only from the original corpus.

\begin{table*}[!ht]
\centering
\caption{\textbf{Comparison with direct retrieval baselines on EnterpriseRAG-Bench.} All rows use the 400-question validation split at the 10M persistent-memory scale. Lexical RAG and single-vector dense RAG retrieve evidence directly from the original corpus, while \method{} uses source-resolved dual-view retrieval. InvDocs is in \([0,1]\), and lower is better. A dash indicates that the baseline does not produce the benchmark-compatible reported-document set required to compute InvDocs.}

\label{tab:headline}
\resizebox{0.95\textwidth}{!}{%
\begin{tabular}{@{}l r r r r r@{}}
\toprule
System & Overall & Correctness & Completeness & Document Recall & InvDocs \\
\midrule
Lexical RAG & 40.27 & 52.75 & 46.95 & 83.71 & -- \\
Single-vector dense RAG & 43.03 & 55.50 & 51.08 & 83.74 & -- \\
Hierarchy-only memory & 60.48 & 73.00 & 64.19 & 68.59 & 0.818 \\
Relation-only memory & 59.16 & 72.50 & 63.02 & 66.74 & 0.828 \\
Published benchmark leader & 68.22 & 81.60 & 72.86 & 79.02 & 0.470 \\
Reranked dual view & 74.46 & 84.25 & 77.58 & 75.78 & 0.804 \\
FullExtractor + Expansion & 82.26 & 86.50 & 86.98 & 81.90 & 0.774 \\
FullExtractor + Expansion + Attribution & 82.26 & 86.50 & 86.98 & 81.90 & 0.760 \\
\bottomrule
\end{tabular}
}

\end{table*}

Post-answer attribution further improves source reporting, reducing InvDocs from \(0.774\) to \(0.760\) without changing answer quality. This confirms that attribution removes unused or unsupported sources while preserving the generated answer.

\subsection{Component Analysis}
\label{sec:ablation}

Table~\ref{tab:ablation} reports a leave-one-component-out study on all 500 EnterpriseRAG-Bench questions at the 10M memory scale. The answer model, embeddings, evaluator, candidate budget, and evidence budget are fixed across all runs; each row removes only the specified component from the full \method{} pipeline.

\begin{table*}[!ht]
\centering
\caption{\textbf{Leave-one-component-out ablation on EnterpriseRAG-Bench.} All rows use the full 500-question benchmark, 10M persistent memory, FullExtractor, GPT-5.4 answering and evaluation, and fixed candidate and evidence budgets. Change is relative to the full configuration.}

\label{tab:ablation}
\resizebox{0.95\textwidth}{!}{%
\begin{tabular}{@{}l r r r r r r@{}}
\toprule
Configuration & Overall & Change (\%) & Correctness & Document Recall & Completeness & Reported Documents \\
\midrule
Full configuration & 82.26 & -- & 86.50 & 81.90 & 86.98 & 2.31 \\
No query expansion & 79.34 & -3.55 & 83.25 & 81.90 & 84.58 & 2.31 \\
No reranker & 78.17 & -4.97 & 82.75 & 81.90 & 83.74 & 2.31 \\
No distillation & 70.62 & -14.15 & 74.75 & 73.10 & 77.64 & 2.31 \\
No dual index & 67.53 & -17.91 & 72.50 & 70.20 & 75.18 & 2.31 \\
No post-answer attribution & 82.26 & 0.00 & 86.50 & 81.90 & 86.98 & 5.00 \\
\bottomrule
\end{tabular}%
}

\end{table*}

The dual index and distillation provide the largest gains. Removing the dual index reduces Overall by \(17.91\%\) and Document Recall from \(81.90\%\) to \(70.20\%\), while removing distillation reduces Overall by \(14.15\%\). Query expansion and reranking provide additional improvements of \(3.55\%\) and \(4.97\%\), respectively. These results show that complementary retrieval representations determine the quality of the candidate set, while query transformation and reranking refine the final evidence selection.

Post-answer attribution serves a different role. Removing it leaves all answer-quality metrics unchanged but increases the average number of reported documents from \(2.31\) to \(5.00\). Thus, attribution removes \(53.8\%\) of the retrieved source set without changing the generated answer. Query expansion also recovers 13 of 33 targets missed by the base retriever, further confirming its contribution to recall.

\subsection{Efficiency and Transfer}
\label{sec:efficiency-transfer}

\paragraph{External transfer.}
\label{sec:external}

Table~\ref{tab:external} evaluates whether \method{} transfers beyond EnterpriseRAG-Bench. On FinanceBench~\citep{islam2023financebench} and HotpotQA~\citep{yang2018hotpotqa}, \method{} achieves Overall scores of 81.05 and 86.37, demonstrating strong performance on financial question answering and multi-hop reasoning. LoCoMo~\citep{maharana2024locomo} and UltraDomain~\citep{qian2024memorag} are more challenging because they require retrieval over broader shared stores with temporal dependencies, cross-session evidence, and cross-domain distractors. Although performance is lower in these settings, the results identify clear directions for extending the current document-oriented memory representation.

\begin{table*}[!ht]
\centering
\caption{\textbf{Cross-dataset transfer results.} FullExtractor and GPT-5.4 are evaluated on each benchmark using its stated question count and task-specific memory setting. N/A indicates that the benchmark does not define the corresponding document metric. InvDocs is in \([0,1]\), and lower is better.}

\label{tab:external}
\resizebox{0.75\linewidth}{!}{%
\begin{tabular}{@{}l r r r r r r@{}}
\toprule
Benchmark & Questions & Correctness & Completeness & Overall & Document Recall & InvDocs \\
\midrule
FinanceBench & 150 & 82.00 & 87.29 & 81.05 & N/A & N/A \\
HotpotQA & 200 & 87.00 & 87.09 & 86.37 & N/A & N/A \\
LoCoMo & 200 & 38.00 & 42.49 & 42.34 & 52.98 & 0.879 \\
UltraDomain & 200 & 9.50 & 13.28 & 17.73 & 45.00 & 0.910 \\
\bottomrule
\end{tabular}%
}

\end{table*}

\paragraph{Evidence efficiency.}
\label{sec:tokens}

We next compare three evidence policies while fixing candidate depth at 20 and sampling 100 questions at each memory scale. As shown in Table~\ref{tab:context}, \method{} retains most of the accuracy of detailed-evidence-only generation while using substantially fewer input tokens. At 10M and 20M, selective detail reduces answer input by \(68.6\%\) and \(68.8\%\), respectively, with only a 1.5-point reduction in Correctness. In contrast, distilled-only evidence is much shorter but loses most of the information required for accurate generation.

\begin{table*}[!ht]
\centering
\caption{\textbf{Evidence-efficiency comparison.} Each row uses 100 sampled questions, candidate depth 20, GPT-5.4 answering, and the stated persistent-memory scale. Token change is measured relative to detailed evidence only.}

\label{tab:context}
\resizebox{0.75\linewidth}{!}{%
\begin{tabular}{@{}r l r r r@{}}
\toprule
Memory scale & Evidence policy & Answer tokens/query & Token change & Correctness \\
\midrule
10M & Detailed evidence only & 6,532 & 0.0\% & 88.00 \\
10M & Selective detail (\method{}) & 2,049 & \(-68.6\%\) & 86.50 \\
10M & Distilled only & 1,013 & \(-84.5\%\) & 21.00 \\
\midrule
20M & Detailed evidence only & 6,497 & 0.0\% & 85.50 \\
20M & Selective detail (\method{}) & 2,025 & \(-68.8\%\) & 84.00 \\
20M & Distilled only & 1,020 & \(-84.3\%\) & 17.00 \\
\bottomrule
\end{tabular}
}

\end{table*}

These results show that \method{} provides a favorable accuracy--context trade-off: it reduces answer-model input by more than two thirds while preserving nearly all of the Correctness obtained with full detailed evidence. This supports the central design choice of using compressed memories for retrieval and loading detailed evidence only when needed.

\paragraph{Operational efficiency.}

The bounded evidence context keeps online inference efficient as persistent memory grows. Building the memory remains an offline cost: constructing 60M tokens requires 9.3 hours, while 250M tokens requires approximately 5.6 days. Retrieval takes only 0.30--1.10 seconds at 60M--100M, end-to-end answering takes 2.5--3.3 seconds, and the complete 500-question diagnostic costs \$3.75, or \$0.0075 per question.

\paragraph{Answerability-aware expansion.}

Query expansion substantially improves retrieval for answerable questions, producing relative gains of 20.8--100.0\% across project, high-level, constrained, conflict, and semantic categories. However, it reduces information-not-found Correctness from 100.0\% to 68.8\%. A selective oracle preserves the gains on answerable categories while restoring information-not-found Correctness to 100.0\%, indicating that a calibrated answerability gate is the main remaining requirement.



\section{Conclusion}

\label{sec:conclusion}

We introduced \method{}, a retrieval solution for ultra-large persistent memory. By using distilled memories for semantic access and resolving all retrieved content to detailed source evidence, \method{} keeps hundreds of millions of tokens searchable while maintaining a bounded generation context. Experiments on EnterpriseRAG-Bench show that this design improves retrieval and answering quality while remaining effective at scales up to 250M tokens. This separation provides a practical foundation for scalable and evidence-grounded persistent memory.

\section*{Limitations}

Most reported configurations are based on single runs, and the detailed-evidence-only packing result is estimated from a retained trace rather than a new controlled rerun. Source precision remains lower than the published EnterpriseRAG-Bench leader, and performance on LoCoMo and UltraDomain shows that the current document-oriented retrieval design does not transfer uniformly to conversational or broad cross-domain memory stores. Finally, \method{} increases the amount of context that can be searched, but it does not extend the native attention window of the underlying language model.

\bibliography{refs}

\clearpage
\appendix

\section{Full Experimental Protocol}
\label{app:protocol}

Table~\ref{tab:protocol} summarizes the complete experimental design, including the question split, memory scale, extractor, evidence setting, answer model, and purpose of each analysis. Because these protocols differ in data partition and system configuration, their absolute values should be interpreted only within the corresponding experiment.

\begin{table*}[!ht]
\centering
\caption{\textbf{Complete experimental protocol.} The 100-question development split is used only for architecture, prompt, and hyperparameter selection. All remaining analyses preserve their own question sets, memory scales, and evaluation purposes.}
\label{tab:protocol}
\resizebox{\textwidth}{!}{%
\begin{tabular}{@{}l l l l l l l@{}}
\toprule
Analysis & Questions & Memory scale & Extraction & Evidence setting & Answer model & Purpose \\
\midrule
Main comparison & 400 validation & 10M & FullExtractor & expansion + attribution & GPT-5.4 & end-to-end quality \\
Scaling study & 400 validation & 20--250M & MiniExtractor & dual-view retrieval & GPT-5.4 & scalability \\
Gold intervention & 400 validation & 20M, 60M, 250M & both & retrieved / gold & GPT-5.4 & failure localization \\
Component ablation & all 500 & 10M & FullExtractor & one component removed & GPT-5.4 & component contribution \\
Evidence efficiency & 100 per scale & 10M, 20M & retained stack & depth 20 & GPT-5.4 & token--quality trade-off \\
Mechanism diagnostics & all 500 & 10M & FullExtractor & varies & GPT-5.4 & retrieval and attribution analysis \\
External transfer & task-specific & task-specific & FullExtractor & task-specific & GPT-5.4 & cross-dataset transfer \\
\bottomrule
\end{tabular}%
}
\end{table*}

The system is developed on 100 questions and then frozen before evaluation on the disjoint 400-question validation split. The full-benchmark ablation, sampled evidence-efficiency study, mechanism diagnostics, and external transfer experiments are reported separately and are not used as substitutes for the main validation results. EnterpriseRAG-Bench metrics follow the benchmark's canonical gold-answer evaluation protocol. Query expansion recovers 13 of 33 targets missed by the base retriever, while post-answer attribution reduces the average reported source set from 5.00 to 2.31 documents.

\paragraph{Backend and vector storage.}
We implement the retrieval backend with disk-persistent ChromaDB stores using \texttt{PersistentClient}, with a separate persistence directory for each memory scale~\citep{chromadb2026client}. The 60M, 100M, and 250M stores contain 49,528, 82,777, and approximately 190K documents, respectively, with the 250M store occupying approximately 60\,GB on disk. Each store contains two independently embedded and queried collections: \texttt{raw\_chunks}, which stores deterministic chunks of approximately 400 \texttt{cl100k\_base} tokens, and \texttt{distilled\_memory}, which stores up to three typed memory entries derived from each chunk. Collections are created with \texttt{get\_or\_create\_collection} and reused in read-only mode during inference~\citep{chromadb2026collections}. Both collections use \texttt{text-embedding-3-small} with 1,536-dimensional vectors~\citep{openai2024embedding3} and ChromaDB's HNSW approximate-nearest-neighbor index with cosine distance. The two collections are queried independently and fused after distilled hits are resolved to their source chunks, implementing the dual-index retrieval component evaluated in our ablation.

\section{Diagnosing Retrieval Degradation with Gold Evidence}
\label{sec:oracle}

We use a gold-evidence intervention to determine whether performance degradation at larger memory scales is caused primarily by retrieval or by answer generation. For each memory scale, GPT-5.4 answers either from the evidence retrieved by \method{} or from the benchmark gold evidence. If generation were the main bottleneck, Correctness would decline similarly in both settings. Instead, performance remains nearly stable with gold evidence but drops substantially with retrieved evidence.

\begin{table*}[!ht]
\centering
\caption{\textbf{Correctness with retrieved and gold evidence across memory scales.} Both extractors use the 400-question validation split and GPT-5.4 answering. Change is measured from 20M to 250M tokens.}
\label{tab:oracle}
\resizebox{0.65\linewidth}{!}{%
\begin{tabular}{@{}l l r r r r@{}}
\toprule
Extractor & Evidence source & 20M & 60M & 250M & Change (\%) \\
\midrule
MiniExtractor & Gold & 82.50 & 84.00 & 82.25 & \(-0.30\) \\
MiniExtractor & Retrieved & 84.00 & 78.25 & 73.50 & \(-12.50\) \\
FullExtractor & Gold & 85.75 & 85.00 & 83.75 & \(-2.33\) \\
FullExtractor & Retrieved & 70.25 & 66.00 & 56.25 & \(-19.93\) \\
\bottomrule
\end{tabular}
}
\end{table*}

\begin{figure}[!ht]
\centering
\includegraphics[width=\columnwidth]{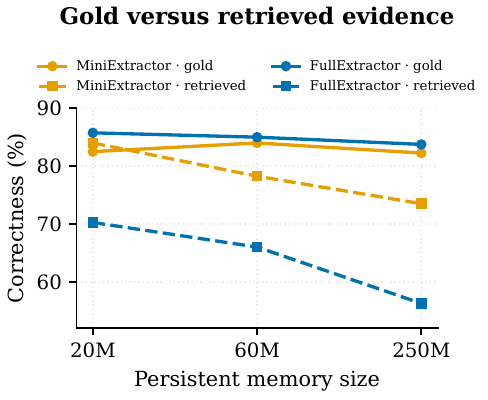}
\caption{\textbf{Separating retrieval failure from generation failure.} Correctness remains stable when GPT-5.4 receives gold evidence, whereas it declines with retrieved evidence as persistent memory grows.}
\label{fig:oracle}
\end{figure}

From 20M to 250M tokens, gold-evidence Correctness changes by only \(0.30\%\) with MiniExtractor and \(2.33\%\) with FullExtractor, compared with \(12.50\%\) and \(19.93\%\) under retrieved evidence. This contrast shows that most of the observed scaling loss arises from retrieving the right evidence from a larger corpus rather than from the answer model's ability to use that evidence.

\section{From Compressed Memory to Source-Resolved Dual-View Retrieval}
\label{app:development}
\label{app:complete-results}

This section summarizes the architecture development that led to the final \method{} design. We first explored hierarchical and relation-based memory structures, which compress the corpus into coarse summaries or semantic links. Although these representations reduced the search space, they did not reliably recover the exact source evidence required for answering. This limitation motivated the shift to dual-view retrieval, where compact memories support semantic access and detailed chunks preserve source-faithful evidence.


Table~\ref{tab:development} summarize the architecture development process. The first two variants organize extracted concept memories through hierarchical or relational structure, while Cognitive Document Memory combines these forms of concept organization, following prior work on long-term conversational memory~\cite{maharana2024locomo}. Dual-view memory then replaces concept-only retrieval with parallel distilled-memory and detailed-evidence indices. Later variants strengthen memory extraction and source resolution, add cross-encoder reranking~\cite{nogueira2019passage}, and introduce query expansion~\cite{wang2023query2doc}. The final attribution stage identifies the loaded sources supporting the fixed answer~\cite{gao2023enabling}. Because several components change between successive rows, these results describe the architecture development trajectory rather than a controlled ablation.

\begin{table*}[!ht]
\centering
\caption{\textbf{Architecture development on EnterpriseRAG-Bench.} Each system is selected using the fixed 100-question development split and evaluated on the disjoint 400-question validation split. The rows record successive architecture changes rather than leave-one-component-out effects. InvDocs is bounded in \([0,1]\), and lower is better.}\label{tab:development}
\resizebox{\textwidth}{!}{%
\begin{tabular}{@{}lrrrrr@{}}
\toprule
System & Overall & Document Recall (\%) & Correctness (\%) & Completeness (\%) & InvDocs \\
\midrule
Hierarchy-only concept memory & 60.48 & 68.59 & 73.00 & 64.19 & 0.818 \\
Relation-only concept memory & 59.16 & 66.74 & 72.50 & 63.02 & 0.828 \\
Cognitive Document Memory & 65.14 & 70.53 & 76.75 & 68.83 & 0.816 \\
Cognitive Document Memory & 66.03 & 70.53 & 78.25 & 69.50 & 0.816 \\
Initial dual-view memory & 67.42 & 74.80 & 78.00 & 71.21 & 0.804 \\
Improved dual-view memory & 72.44 & 76.03 & 83.25 & 76.35 & 0.802 \\
Reranked dual-view memory & 74.46 & 75.78 & 84.25 & 77.58 & 0.804 \\
Expanded-control branch & 66.40 & 74.69 & 85.50 & 69.63 & 0.806 \\
FullExtractor + Expansion & 82.26 & 81.90 & 86.50 & 86.98 & 0.774 \\
FullExtractor + Expansion + Attribution & 82.26 & 81.90 & 86.50 & 86.98 & 0.760 \\
\bottomrule
\end{tabular}%
}
\end{table*}

\begin{figure}[!ht]
\centering
\includegraphics[width=\columnwidth]{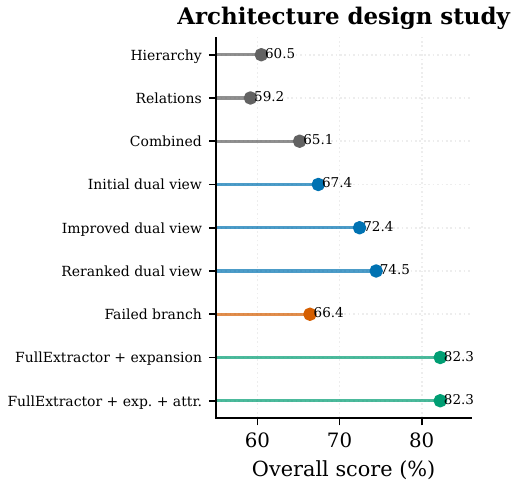}
\caption{\textbf{Performance across successive architecture designs.} Dual-view retrieval provides the main improvement over compressed-only memory, while reranking, query expansion, and attribution further refine evidence selection and source reporting.}
\label{fig:development}
\end{figure}

\paragraph{Development stages.}
Hierarchy-only and relation-only memory use compressed representations as the primary access structures. The initial dual-view system introduces separate detailed-evidence and typed-memory indices, while the improved variant refines source resolution and evidence construction. The reranked dual-view system adds cross-encoder ordering, and FullExtractor with query expansion further improves semantic recall. Post-answer attribution is added last and improves source reporting without changing answer quality. Overall, the development path shows that the key architectural change is the transition from compressed-only memory to source-resolved dual-view retrieval.

\section{Offline Cost, Online Latency, and Evidence-Context Efficiency}
\label{app:efficiency}

This section analyzes the operational cost of \method{} and the trade-off between evidence-context size and answer quality. Table~\ref{tab:operations} separates one-time memory construction from online retrieval, answering, and evaluation.

\begin{table}[!ht]
\centering
\caption{\textbf{Offline construction, online serving, and evaluation cost.} Build times are reported for 60M and 250M tokens, while serving latency is measured at 60M and 100M.}
\label{tab:operations}
\resizebox{\columnwidth}{!}{%
\begin{tabular}{@{}l l r@{}}
\toprule
Regime & Measurement & Value \\
\midrule
Offline construction & 60M memory & 9.3 h \\
Offline construction & 250M memory & approximately 5.6 d \\
Online retrieval & 60M / 100M & 0.30 s / 1.10 s \\
End-to-end answering & 60M / 100M & 2.5 s / 3.3 s \\
Evaluation & 500 questions / per question & \$3.75 / \$0.0075 \\
\bottomrule
\end{tabular}
}
\end{table}

Although memory construction becomes more expensive as the corpus grows, it is performed offline. Online retrieval remains below 1.1 seconds at 100M tokens, and end-to-end answering remains below 3.3 seconds, showing that the bounded evidence context keeps inference practical.

We next compare distilled-only evidence, selective detailed evidence used by \method{}, and full detailed evidence. All settings use the same 100 questions, candidate depth of 20, and GPT-5.4 answer model.

\begin{table*}[!ht]
\centering
\caption{\textbf{Answer quality and context size under different evidence policies.} Token counts are measured per question at 10M and 20M persistent-memory scales.}
\label{tab:context-exact}
\resizebox{0.8\linewidth}{!}{%
\begin{tabular}{@{}l rr rr@{}}
\toprule
& \multicolumn{2}{c}{10M memory} & \multicolumn{2}{c}{20M memory} \\
\cmidrule(lr){2-3}\cmidrule(lr){4-5}
Evidence policy & Tokens/query & Correctness & Tokens/query & Correctness \\
\midrule
Distilled only & 1,013 & 21.00 & 1,020 & 17.00 \\
Selective detail (\method{}) & 2,049 & 86.50 & 2,025 & 84.00 \\
Detailed evidence only & 6,532 & 88.00 & 6,497 & 85.50 \\
\midrule
Detailed/\method{} token ratio & \multicolumn{2}{c}{3.19\(\times\)} & \multicolumn{2}{c}{3.21\(\times\)} \\
\bottomrule
\end{tabular}
}
\end{table*}

\begin{figure}[!ht]
\centering
\includegraphics[width=\columnwidth]{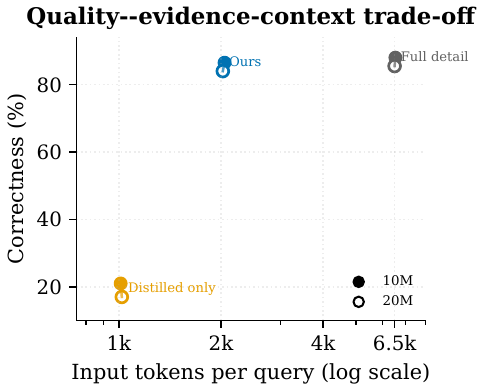}
\caption{\textbf{Accuracy--context trade-off across evidence policies.} \method{} preserves nearly all of the Correctness of full detailed evidence while using approximately one third of the answer-model input.}
\label{fig:context}
\end{figure}

At both memory scales, \method{} reduces the answer context by approximately \(69\%\) relative to full detailed evidence, while sacrificing only 1.5 Correctness points. Distilled-only evidence is shorter but performs substantially worse, confirming that compressed memories are effective retrieval keys but insufficient as generation evidence.

Table~\ref{tab:token-breakdown} reports the per-query workload for the full 500-question diagnostic. Questions are short on average, while answer length is more variable; the system retrieves approximately 34 chunks before reranking and evidence packing.

\begin{table}[!ht]
\centering
\small
\caption{\textbf{Per-query workload under the 500-question diagnostic protocol.}}
\label{tab:token-breakdown}
\begin{tabular}{@{}lrrr@{}}
\toprule
Component & Mean & Median & P95 \\
\midrule
Question input tokens & 37 & 34 & 63 \\
Answer output tokens & 439 & 221 & 1,622 \\
Retrieved chunks & 34.3 & 35 & -- \\
\bottomrule
\end{tabular}
\end{table}

\section{Cross-Dataset Transfer and Query-Expansion Diagnostics}
\label{app:transfer-analysis}

This section analyzes two complementary questions: how well \method{} transfers across retrieval regimes, and how query expansion affects recall and abstention.

\begin{figure}[!ht]
\centering
\includegraphics[width=\columnwidth]{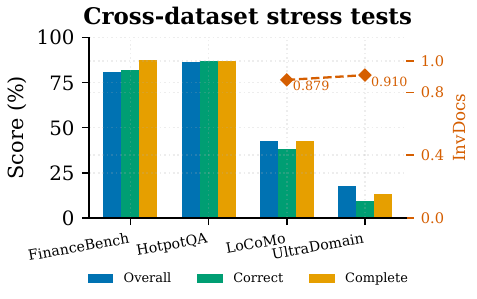}
\caption{\textbf{Transfer performance across evidence regimes.} FullExtractor and GPT-5.4 are evaluated on 150 FinanceBench questions and 200 questions for each remaining benchmark. Document-level metrics are undefined for FinanceBench and HotpotQA.}
\label{fig:external}
\end{figure}

\paragraph{Cross-dataset transfer.}
Figure~\ref{fig:external} visualizes the results in Table~\ref{tab:external}. \method{} performs strongly on FinanceBench and HotpotQA, where evidence is provided within a question-local corpus, but is less effective on LoCoMo and UltraDomain, which require retrieval over broader conversational or cross-domain stores. This contrast indicates that the current document-oriented dual-view index transfers best when evidence units and corpus structure are well aligned with its memory representation.

\paragraph{Retrieval and attribution effects.}
Query expansion recovers 13 of 33 targets missed by the base retriever, corresponding to a 39.4\% recovery rate. Post-answer attribution reduces the average reported source set from 5.00 to 2.31 documents, removing 53.8\% of retrieved sources without changing the answer.

\begin{table}[!ht]
\centering
\caption{\textbf{Mechanism-level retrieval and attribution diagnostics.} Both analyses use all 500 EnterpriseRAG-Bench questions at 10M with FullExtractor.}
\label{tab:gates}
\resizebox{\columnwidth}{!}{%
\begin{tabular}{@{}l r@{}}
\toprule
Diagnostic & Result \\
\midrule
Query-expansion target recovery & 13/33 (39.4\%) \\
Evidence pool \(\rightarrow\) reported documents & 5.00 \(\rightarrow\) 2.31 (53.8\% removed) \\
\bottomrule
\end{tabular}
}
\end{table}

\paragraph{Query expansion and answerability.}
Table~\ref{tab:type-analysis} shows that query expansion improves Correctness across all five answerable question categories, with relative gains ranging from 20.8\% to 100.0\%. The largest gains appear on project-related and high-level questions, where terminology mismatch and indirect references make direct retrieval difficult.

\begin{table}[!ht]
\centering
\caption{\textbf{Correctness by question type before and after query expansion.} All 500 EnterpriseRAG-Bench questions are evaluated at 10M with FullExtractor.}
\label{tab:type-analysis}
\setlength{\tabcolsep}{3pt}
\resizebox{\columnwidth}{!}{%
\begin{tabular}{@{}l r r r@{}}
\toprule
Question type & Pre-exp. (\%) & Unconditional (\%) & Selective oracle (\%) \\
\midrule
Project-related & 53.10 & 96.90 & 96.90 \\
High-level & 37.50 & 75.00 & 75.00 \\
Constrained & 66.70 & 87.50 & 87.50 \\
Conflict & 81.20 & 100.00 & 100.00 \\
Semantic & 72.00 & 87.00 & 87.00 \\
Info not found & 100.00 & 68.80 & 100.00 \\
\bottomrule
\end{tabular}
}
\end{table}

The main trade-off appears on information-not-found questions. Unconditional expansion answers five of the 16 unanswerable cases, reducing Correctness from 100.0\% to 68.8\%. A selective oracle preserves all gains on answerable questions while restoring information-not-found Correctness to 100.0\%. This result shows that query expansion and reliable abstention are compatible, but require an answerability-aware gate that decides when expansion should be applied.

\section{Memory Construction, Inference Order, and Reproducibility}
\label{app:implementation}

\paragraph{Dual-view memory construction.}
Algorithm~\ref{alg:build} gives the complete construction procedure. Each distilled memory is stored with the immutable identifier of the detailed source chunk from which it is derived, ensuring that both indices refer to the same underlying evidence.

\begin{algorithm}[!ht]
\caption{Construct source-resolved dual-view memory}
\label{alg:build}
\small
\begin{algorithmic}[1]
\Require documents \(\mathcal{D}\), memory extractor \(\mathcal{E}_{m}\), embedding function \(\phi\)
\ForAll{detailed source chunks \(x_i\) in \(\mathcal{D}\)}
    \State add \((i,x_i,\phi(x_i))\) to the detailed index
    \ForAll{distilled memory \(m\in\mathcal{E}_{m}(x_i)\)}
        \State add \((m,\phi(m),\pi(m)=i)\) to the distilled index
    \EndFor
\EndFor
\State \Return detailed index, distilled index, and source map \(\pi\)
\end{algorithmic}
\end{algorithm}

\paragraph{Inference and attribution.}
Figure~\ref{fig:overview} summarizes the online pipeline. Original and transformed queries first retrieve candidates from both indices. Distilled hits are then resolved to detailed source chunks before fusion, reranking, and evidence packing. The answer is generated from the resulting bounded evidence context, after which attribution selects only the loaded sources that support the fixed answer.


\begin{algorithm}[!ht]
\caption{Generate an answer and attribute supporting sources}
\label{alg:answer}
\small
\begin{algorithmic}[1]
\Require question \(q\), bounded detailed evidence \(E\)
\State \(a\gets\Call{EvidenceOnlyAnswer}{q,E}\)
\State \(Z'\gets\Call{Attribute}{a,E}\)
\State \(Z\gets Z'\cap\operatorname{docs}(E)\)
\State \Return answer \(a\) and supporting sources \(Z\)
\end{algorithmic}
\end{algorithm}

\paragraph{Reproducibility settings.}
All generative calls use temperature \(0\), and data splitting, sampling, and tie-breaking use seed 42. Each request has a 120-second timeout and at most two retries with provider backoff. The evidence context is capped at 4,096 tokens, the generated answer at 800 tokens, and the reranked candidate set at five unique documents. RRF uses \(\gamma=60\), while the context-packing study fixes the first-stage candidate depth at 20.

\paragraph{Models and failure handling.}
Atomic memory extraction uses \texttt{gpt-5.4-mini}~\citep{openai2026gpt54mini}; higher-level abstraction, query transformation, answering, post-answer attribution, and EnterpriseRAG evaluation use \texttt{gpt-5.4}~\citep{openai2026gpt54}; and all indices use \texttt{text-embedding-3-small}~\citep{openai2024embedding3}. Extraction, query transformation, and attribution return schema-validated JSON. After all retries fail, extraction returns no memories, query transformation falls back to the original question, attribution returns an empty set, and answering returns the predefined abstention response. Parsed citations are finally intersected with the document identifiers included in the evidence context, preventing unsupported sources from being reported.

\section{Visualization of Main Results and Diagnostics}
\label{app:visualizations}

This section visualizes the main system comparison, component ablation, query-expansion behavior, and token usage. Exact values are reported in the corresponding tables.

\begin{figure}[!ht]
\centering
\includegraphics[width=\columnwidth]{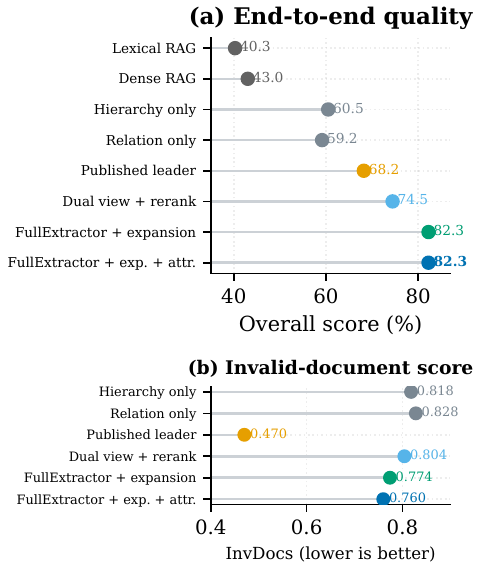}
\caption{\textbf{End-to-end performance at the 10M memory scale.} The figure summarizes the 400-question validation results in Table~\ref{tab:headline}, comparing direct retrieval baselines, compressed-memory variants, and \method{}. InvDocs is in \([0,1]\), and lower is better.}
\label{fig:headline}
\end{figure}

Figure~\ref{fig:headline} summarizes end-to-end performance at the 10M memory scale. \method{} achieves the strongest overall result among direct retrieval baselines and compressed-memory variants, while also maintaining competitive document recall and a lower invalid-document ratio.

\begin{figure}[!ht]
\centering
\includegraphics[width=\columnwidth]{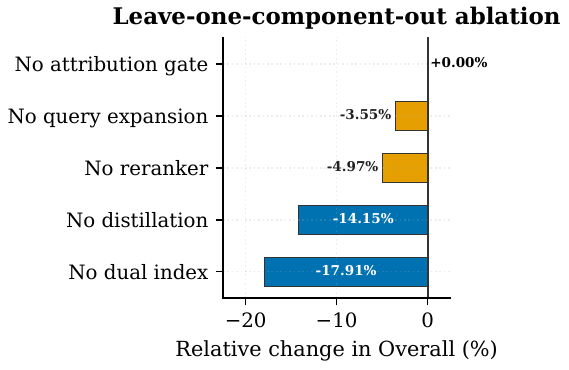}
\caption{\textbf{Contribution of individual \method{} components.} The figure shows the relative change in Overall after removing each component from the full system under the 500-question ablation protocol.}
\label{fig:ablation}
\end{figure}

Figure~\ref{fig:ablation} shows how each component contributes to the full system. Removing any major component reduces Overall, confirming that dual-view memory, reranking, query expansion, and attribution support complementary parts of the final pipeline.

\begin{figure}[!ht]
\centering
\includegraphics[width=\columnwidth]{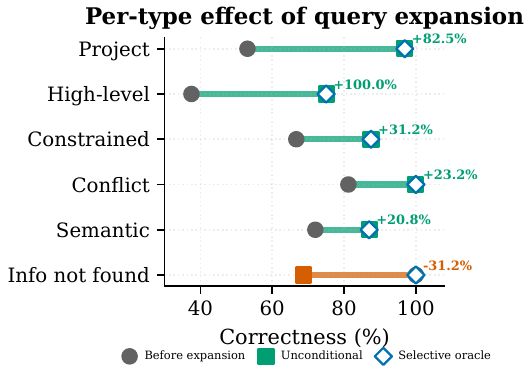}
\caption{\textbf{Effect of query expansion across question types.} Query expansion improves all answerable categories, while an answerability-aware gate is required to preserve performance on information-not-found questions.}
\label{fig:type-analysis}
\end{figure}

Figure~\ref{fig:type-analysis} examines how query expansion affects different question types. Expansion improves performance across answerable categories, but information-not-found questions require an answerability-aware gate to prevent expanded queries from retrieving plausible but unsupported evidence.

\section{Why Dual-View Retrieval Helps and Where Scaling Fails}
\label{app:analysis}

This section provides a simple interpretation of the two main empirical findings: dual-view retrieval improves evidence recall, while performance degradation at larger memory scales is driven mainly by retrieval.

Let \(A\) denote the event that relevant evidence is retrieved from the detailed index, \(M\) the event that it is retrieved from the distilled index, and \(G\) the event that a distilled hit is correctly resolved to its detailed source chunk. The recall of dual-view retrieval is
\begin{equation}
\begin{split}
R_{\mathrm{dual}}
&=\Pr\left(A\cup(M\cap G)\right)\\
&=R_A+\Pr(M\cap G\cap A^c),
\end{split}
\label{eq:dual-recall}
\end{equation}
where the second term captures evidence recovered by the distilled route but missed by the detailed route. This term explains the complementarity observed in the ablation study.

To interpret scaling behavior, let \(S_N\) denote the event that sufficient evidence is selected from a corpus of size \(N\). Answer Correctness can be written as
\begin{align}
\Pr(Y=1\mid N)&=\Pr(S_N)\Pr(Y=1\mid S_N)\\
&+\Pr(\neg S_N)\Pr(Y=1\mid\neg S_N).
\label{eq:decompose}
\end{align}
As memory grows, the main changing term is \(\Pr(S_N)\), since relevant evidence must be identified among more distractors. The gold-evidence experiment approximately controls for evidence sufficiency and shows that Correctness remains stable once the required evidence is provided. This supports the conclusion that the observed scaling loss is primarily caused by retrieval rather than answer generation.

\section{Prompt Templates}
\label{app:prompts}

This section provides the prompt templates used for memory construction, query transformation, candidate selection, answer generation, attribution, and evaluation. The prompts are reproduced in their original form for completeness and reproducibility.

\begin{promptbox}[teal]{Atomic typed-memory extraction}
\textbf{System:} Extract at most three atomic, retrieval-friendly memories entailed by the
source chunk.  Each memory must have a type from
\{fact, procedure, definition, requirement, decision\}, a short retrieval key, a concise
value, and no outside knowledge.  Skip filler.  Return
\texttt{\{"memories":[...]\}}; return an empty list when the chunk has no useful content.

\textbf{User:} Document chunk, preceded by its section path and immutable chunk identifier.
\end{promptbox}

\begin{promptbox}[blue]{High-level memory abstraction}
\textbf{System:} Summarize the supplied typed memories into a compact search key for their
shared topic.  Preserve named entities, constraints, dates, exceptions, and conflicts.  Do
not create a fact absent from the children.  Return JSON containing
\texttt{search\_key}, \texttt{summary}, and the unchanged list of child source identifiers.

\textbf{User:} A bounded group of typed memories with their immutable document mappings.
\end{promptbox}

\begin{promptbox}[orange]{Query transformation}
\textbf{System:} Preserve the user's information need and named constraints.  Produce a
canonical rewrite and terminology-diverse alternatives; do not invent an answer, entity, or
date.  Return a JSON list.  If no safe reformulation exists, return only the original query.

\textbf{User:} The original question.
\end{promptbox}

\begin{promptbox}[violet]{Cross-route candidate selection}
\textbf{System:} Select only candidate identifiers that could answer the question under its
stated entities, time range, and constraints.  A typed-memory match is a search clue, not
evidence; retain its linked detailed-evidence identifier.  Return ranked identifiers and short selection
reasons in JSON.  Never invent or rewrite an identifier.

\textbf{User:} Question, fused detailed and typed candidates, route ranks, and source metadata.
\end{promptbox}

\begin{promptbox}[red]{Evidence-only answer}
\textbf{System:} Answer using only the supplied document evidence.  Preserve qualifiers and
conflicts.  If the evidence is insufficient, reply exactly:
\texttt{I don't have enough information to answer.}

\textbf{User:} The question followed by evidence cards labeled with chunk identifier,
document identifier, and section path.
\end{promptbox}

\begin{promptbox}[green!60!black]{Post-answer attribution}
\textbf{System:} The answer is fixed.  Return only identifiers of supplied evidence cards
that directly support a claim in that answer.  Do not add a source, revise the answer, or cite
topically related but unused evidence.  Return \texttt{\{"document\_ids":[...]\}}.

\textbf{User:} Fixed answer plus the same evidence cards shown to the answer model.
\end{promptbox}

\begin{promptbox}[magenta]{EnterpriseRAG correctness evaluation}
\textbf{System:} Compare the candidate answer with the benchmark gold answer.  Judge whether
the candidate preserves the required conclusion, scope, conditions, and conflicts.  Do not
reward unsupported details.  Return schema-constrained JSON with
\texttt{correct}, \texttt{validated\_facts}, and \texttt{missing\_facts}; do not expose or
use any field outside the canonical gold-answer protocol.

\textbf{User:} Question, candidate answer, and canonical gold answer.
\end{promptbox}

\end{document}